# A Vision for Operationalising Diversity and Inclusion in AI

Muneera Bano
CSIRO's Data61
Clayton, VIC, Australia
muneera.bano@csiro.au

Didar Zowghi
CSIRO's Data61
Eveleigh, NSW, Australia
didar.zowghi@csiro.au

Vincenzo Gervasi
Department of Computer Science
Pisa University, Italy
vincenzo.gervasi@unipi.it

## ABSTRACT

The growing presence of Artificial Intelligence (AI) in various sectors necessitates systems that accurately reflect societal diversity. This study seeks to envision the operationalization of the ethical imperatives of diversity and inclusion (D&I) within AI ecosystems, addressing the current disconnect between ethical guidelines and their practical implementation. A significant challenge in AI development is the effective operationalization of D&I principles, which is critical to prevent the reinforcement of existing biases and ensure equity across AI applications. This paper proposes a vision of a framework for developing a tool utilizing persona-based simulation by Generative AI (GenAI). The approach aims to facilitate the representation of the needs of diverse users in the requirements analysis process for AI software. The proposed framework is expected to lead to a comprehensive persona repository with diverse attributes that inform the development process with detailed user narratives. This research contributes to the development of an inclusive AI paradigm that ensures future technological advances are designed with a commitment to the diverse fabric of humanity.

## CCS CONCEPTS

• Social and professional topics ~ User characteristics • Software and its engineering ~ Software creation and management ~ Designing software ~ Requirements analysis • Computing methodologies ~ Artificial intelligence ~ Natural language processing ~ Natural language generation

## 1. Introduction

The pervasive role of Artificial Intelligence (AI) in social interactions, from generating and recommending content to processing images and voices, brings numerous benefits but also necessitates addressing ethical implications and risks, such as ensuring equitable and non-discriminatory decision-making, and preventing the amplification of existing inequalities and biases [1]. Diversity and inclusion (D&I) in AI involves considering differences and underrepresented perspectives in AI development and deployment while addressing potential biases and promoting equitable outcomes for all stakeholders [1]. Incorporating D&I principles in AI can enable technology to better respond to the needs of diverse users while upholding ethical values of fairness, transparency, and accountability [2].

Requirements engineering (RE) is an essential part of software development in general, and in developing AI systems in particular. RE includes the identification, analysis, and specification of stakeholder needs, ideally captured in a consistent and precise manner. By focusing on users' and stakeholders' needs, RE aims to contribute to achieving user satisfaction and successful system adoption [3]. Using traditional RE practices for AI systems presents new challenges [4], as these traditional methods need to evolve to address somewhat different AI systems requirements, including those related to data and ethics [5, 6].

To develop ethical and trustworthy AI systems, it is recommended to consider embedding D&I principles throughout the entire development and deployment lifecycle [7]. Overlooking D&I aspects can result in issues related to fairness, trust, bias, and transparency, potentially leading to digital redlining, discrimination, and algorithmic oppression [8]. We posit that RE processes and practices can be tailored and adopted to identify and analyse AI risks and navigate trade-offs and conflicts that may arise due to neglecting D&I principles. For instance, RE can facilitate decision-making in scenarios where maximising inclusion may negatively affect performance or efficiency or when data transparency about members of under-represented groups could compromise their privacy. By acknowledging diverse users and emphasising inclusive system development, RE can aid in achieving a balance between conflicting objectives and promote the development of ethical AI systems [9].

Embracing diversity and inclusion (D&I) is not just an ethical imperative but a strategic advantage in today's global market. For organizations in the software development sector, the stakes are particularly high. The software they create serves as the backbone for communities that span the spectrum of human experience, including those historically marginalized or with unique needs. The challenge of incorporating D&I into both corporate culture and product design is significant.

The emergence of AI technologies within software systems has intensified the urgency for D&I. These systems, often opaque in their workings, can inadvertently perpetuate biases, leaving certain users feeling overlooked or unfairly treated. It is insufficient to cater only to the most prevalent user demographics. The inclusion of distinct voices, especially those defined by so-called protected attributes like gender, race, religion, or ethnicity, is critical. Everyone is unique and may be identified with a number of these attributes, and the multiplicity of their combinations is vast.

However, resource limitations typically constrain efforts to engage with this diversity during the requirements elicitation phase. This often results in a disproportionate focus on larger represented groups or those protected by law, like individuals with disabilities. Such limitations can lead to the exclusion of other equally important perspectives, with the resulting AI software failing to serve a truly diverse user base. Moreover, even when diverse stakeholders are included, there's no assurance that their input will represent the breadth of experiences within their demographic. While larger groups may provide a more substantial pool for elicitation, ensuring a genuinely representative consensus remains a challenge.

There is an urgent need for methodologies and associated tools that enable software development organisations, especially those developing and working with AI components, to integrate a wide array of viewpoints into their RE processes. These methodologies must be:

- **Economically viable**, considering the finite resources available for D&I without compromising other development priorities.
- **Comprehensive in stakeholder representation**, ensuring that the system's impact on diverse user groups is considered.
- **Traceable**, allowing developers to demonstrate which D&I concerns were addressed and how they influenced the design choices.

Addressing these requirements is not merely a necessity but a moral and business imperative, ensuring that the AI systems are equitable and inclusive.

The rest of the paper is structured as follows: Section 2 introduces the concept of D&I in AI. Section 3 gives an overview of existing research on Personas, within RE, advances with large language models and within the space of D&I. Section 4 presents research motivation and section 5 describes the vision of the research in detail. Section 6 presents illustrative case studies and section 7 describes expected results. Section 8 concludes the paper.

## 2. Diversity and Inclusion in AI

The concept of D&I in AI is increasingly recognized as paramount, yet there remains a notable void in translating these principles into tangible practices within AI systems. Fosch-Villaronga and Poulsen [10] discuss D&I in AI as encompassing both the technical fabric of AI and its socio-cultural footprint, advocating for a representation that traverses the spectrum of socio-political power differentials, such as those defined by gender and race. Inclusion is conceptualized as the degree to which individual users find the AI system's offerings relevant and accessible to them. This multifaceted notion of D&I is dissected across three levels: the technical, ensuring algorithms are comprehensive and non-discriminatory; the community, advocating for diversity within AI development teams; and the user, focusing on the stakeholders' involvement and feedback in the AI endeavour, in alignment with Responsible Research and Innovation principles.

Despite the growing consensus on the importance of these concepts, the existing literature often lacks a concrete definition and methodical approach for embedding D&I into AI. In response to this gap, Zowghi and da Rimini [11] proposed a comprehensive definition and a framework to guide the integration of these principles throughout the AI development lifecycle. This definition, iteratively refined with feedback from Responsible AI and D&I experts [11], adopts a socio-technological perspective. It emphasizes that countering bias and fostering fairness necessitates a comprehensive strategy that accounts for cultural dynamics, actively involves stakeholders, and addresses the intersectionality of user attributes. Their proposed five pillars of D&I in the AI ecosystem (humans, data, process, system, and governance), serve as the foundation for structuring and operationalizing D&I within the AI ecosystem, which also includes the application domain or environment [12].

The integration of AI systems into daily life amplifies the imperative for operationalizing AI ethics, particularly those centring on D&I. Ethical AI guidelines, while abundant, often remain abstract, underscoring the necessity for concrete, actionable methods to ensure AI systems are truly ethical [13-15]. The absence of such guidance threatens to perpetuate existing social biases and marginalization, underscoring the need for consistency in the interpretation and application of ethical principles [14]. The prevalent disconnect between high-level ethical aspirations and their enactment within the AI ecosystem—often muddied by "ethics washing" and corporate opacity—further exacerbates the challenge [16, 17].

The discourse on AI ethics, marked by critiques of the inefficacy of current guidelines [18] and counterarguments advocating for ethical trade-offs [19], underscores a division on how best to transform ethical considerations into actionable development criteria. This research posits that operationalizing D&I requirements in AI is essential to bridge the divide between ethical intent and practice [19].

By approaching D&I from a Requirements Engineering perspective, this research envisions creating a structured, actionable framework that ensures AI systems are developed and deployed with a commitment to inclusivity, equity, and ethics [20-22]. The goal is to move beyond the rhetoric of high-level guidelines and establish a set of best practices that manifest D&I principles in the operational realities of AI systems development.

## 3. Personas

The application of personas in user-centred design has been explored in many areas, ranging from software development, and RE, to marketing, more recently, into the complexities of AI. This interdisciplinary adoption highlights the significance of personas as tools for understanding and integrating the user perspective in product development.

Since Cooper's seminal introduction of the persona technique, it has become a mainstay in design disciplines [23]. The literature brims with accounts of personas enabling development teams to envision the needs and behaviours of their end-users through constructed, narrative-driven character sketches. This practice serves to focus the design process on user experiences, fostering empathy and strategic alignment within development teams [23, 24]. However, while the narrative richness of personas aids in internal communication and design ideation, the literature also cautions against potential pitfalls. These include the oversimplification of user groups and the reinforcement of stereotypes, which can skew design decisions and alienate users [25-28].

### 3.1 Personas in RE

In RE, the integration of personas has been lauded for offering concrete and relatable user stories that guide system requirements [29]. Research highlights how personas can encapsulate complex user interactions with systems, aiding teams in navigating the multifaceted landscape of user needs [30]. Yet, despite their value in RE, personas are not without their risks. Studies point to challenges in ensuring the representativeness of personas and in extending their application beyond surface-level user attributes to include deeper, often unspoken needs [31, 32].

### 3.2 Personas and LLMs

The intersection of personas and Large Language Models (LLMs) marks a recent frontier in persona-related research. LLMs like GPT-3 and GPT-4 have the potential to dynamically generate persona narratives, leveraging vast datasets to produce diverse and nuanced user archetypes [26-28]. This synergy promises efficiency and depth in persona creation but is also accompanied by risks. The literature underscores the tendency of LLMs to perpetuate and even amplify stereotypes, necessitating careful oversight and ethical consideration in their use [26, 33].

### 3.3 Personas and Diversity and Inclusion in AI

To date, there is a scarcity of research dedicated to the utilization of LLMs and personas together that explicitly supports diversity and inclusion principles. The benefits of deploying AI to generate virtual personas could be manifold, particularly in enhancing D&I within AI systems. LLMs capability to simulate a broad range of user personas can lead to more inclusive and accessible products. It also presents a solution to practical constraints such as limited resources, the need for scalability, and the challenges of recruiting diverse participant pools for requirements analysis, and traceability. However, the creation of virtual personas must be approached with caution to avoid the entrenchment of stereotypes and biases.

The literature review in Section 2 points to a pressing need for tools that can provide environments for the generation and simulation of virtual personas. Such tools would enable the exploration of diverse user scenarios without the traditional constraints of persona development, thereby broadening the scope and impact of D&I efforts in AI systems design.

## 4. Research Motivation

The opaque nature of many AI-based systems exacerbates the risk of neglecting the nuanced needs of diverse user groups, potentially leading to a heightened sense of exclusion or discrimination. In response to the rapidly evolving technological milieu and the intricacies of social sensitivities, there is a growing consensus that addressing the needs and values of only the most prevalent user demographics is insufficient. A more granular approach is required, one that accounts for the specific perspectives of distinct groups and individuals, particularly those characterized by protected attributes such as gender, race, religion, and ethnicity.

Given the vast array of potential user profiles that arise from the intersection of these attributes, the traditional methodologies for requirements elicitation prove insufficient. They often fail to capture the full spectrum of stakeholder viewpoints, particularly from smaller or legally undefined minority groups. This limitation not only restricts the inclusivity of the design process but may also result in the perpetuation of systemic biases in the implementation, deployment, and operation of AI-based systems.

To address these challenges, this research posits the need for innovative, affordable, and scalable techniques that enable the consideration of a sufficiently diverse set of viewpoints within the requirements engineering process. Such techniques should offer representativeness without overwhelming the development process with resource demands. Additionally, they must provide traceability, enabling developers to document and justify how D&I concerns have informed design decisions. The use of personas in this context emerges as a promising avenue, offering a cost-effective and representative means to simulate a wide array of stakeholder inputs and facilitate the elicitation of diverse requirements. This further motivates the research to explore and validate the use of personas as a tool for eliciting D&I requirements for AI systems, ensuring that they are implemented thoughtfully and effectively to support the creation of more equitable and inclusive AI.

## 5. Proposed Vision

Our multi-stage research vision aims to innovate the way personas are crafted and utilized in the software development lifecycle, particularly in the context of diversity and inclusion. The goal is to develop a robust, dynamic, and ethically grounded framework that leverages the power of AI and data analytics while ensuring the representation of diverse user groups. The envisioned stages of research are as follows (as summarized in Fig 1):

**Step 1: Creation of a Persona Repository:** The initial stage focuses on the development of a comprehensive persona repository. This repository could be an amalgamation of data from diverse sources such as general census data, and social media inputs as well as specifically generated for the use case by relevant stakeholders from the project sponsors' organsiation. By harnessing the breadth of data available from these sources, we aim to construct a rich, multi-dimensional repository of persona profiles. This repository will not be static; an update loop will be integrated to ensure the persona attributes evolve with societal trends and data availability, maintaining the relevance and accuracy of the personas over time.

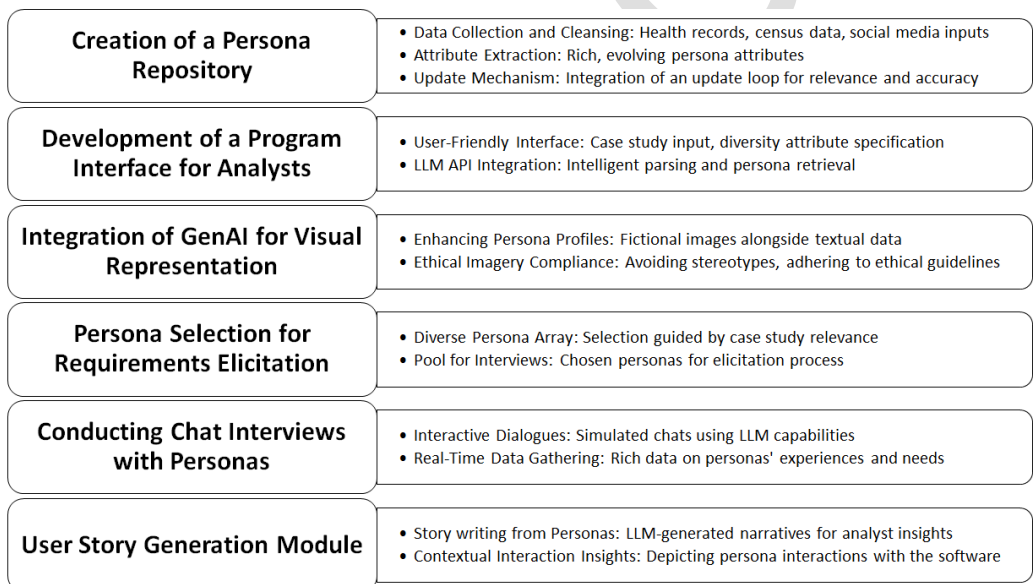

**Figure 1: Vision for D&I in AI requirements elicitation with virtual Personas**

**Step 2: Development of a Program Interface for Analysts:** To operationalize the use of this repository, we propose the creation of a user-friendly program interface for analysts. This interface will act as a gateway for analysts to input their case studies and specify the potential diversity of attributes relevant to their research needs. There could be a potential methodological concern that the analysts might focus only on expected cases, hence, it is crucial to integrate a balanced approach in the operationalization of the proposed repository. The user-interface design would allow analysts to input their case studies and specify relevant diversity attributes and will be enhanced by a LLM's API in the backend. The key to avoiding the trap of encountering only expected cases lies in the LLM's ability to not just retrieve personas matching the specified attributes, but also to intelligently suggest additional, unexpected personas that could provide valuable insights. This functionality ensures that while analysts have the autonomy to define their research parameters, they are concurrently exposed to a wider array of perspectives. Such an approach aligns with the original intent of the repository - to unearth valuable, unforeseen feedback, thus enriching the research process. The integration of this feature in the interface is pivotal, as it fosters a comprehensive exploration that combines both guided and open-ended discovery, crucial for the depth and breadth of research supported by the repository.

In this research framework, the known-unknown matrix is a tool that can categorize the levels of certainty and uncertainty in case studies, aiding analysts in selecting appropriate methods for persona interviews. It distinguishes between known-knowns (cases with clear, defined contexts), unknown-unknowns (cases with high levels of uncertainty), and known-unknowns (cases with a mix of known elements and uncertainties), guiding the use of the repository in a way that optimally aligns with each scenario's specific needs. When dealing with known-known contexts, such as a focus on a particular disability, analysts can define their own diversity attributes for simulated persona interviews, tailoring the selection to specific needs. Conversely, in situations characterized by high uncertainty or unknown-unknowns, the entire repository of personas can be utilized. Here, conducting brief interviews across a broad spectrum with the same questions allows for the identification of recurring themes of diversity and inclusion requirements from specific persona types. For cases falling into the known-unknown category, analysts have the flexibility to employ both techniques, blending targeted exploration with wider, more general inquiry to comprehensively address the varying levels of knowns and unknowns in their research.

**Step 3: Integration of Generative AI for Visual Persona Representation:** Alongside the textual persona profiles, Generative AI (GenAI) will be employed to create corresponding fictional images. The integration of GenAI for visual persona representation in eliciting diversity and inclusion requirements offers significant advantages. By creating visual representations to accompany textual persona profiles, GenAI adds a layer of tangibility and relatability that is inherently more engaging and impactful than text alone. Humans are visual creatures, and the presence of imagery can foster a deeper understanding and empathy towards different personas. This approach is particularly effective in representing a wide spectrum of diversity, allowing for a more inclusive exploration of personas from various backgrounds, abilities, and identities. Crucially, with careful adherence to ethical and diversity & inclusion guidelines, these visual representations can be designed to avoid perpetuating stereotypes, ensuring fair and accurate depictions. This visual element not only enhances engagement but also stimulates creative thinking and insight, often revealing nuances and evoking considerations that text alone may not.

**Step 4: Persona Selection for Requirements Elicitation:** As the next step, the analyst is presented with a carefully curated selection of personas, each representing different facets of the target user base, tailored to the specific nature of the case study using the known-unknown matrix discussed earlier. This strategic selection is crucial for ensuring comprehensive coverage of diversity and aligns with the principles of inclusivity. The selected personas then form the pool for subsequent requirements elicitation interviews, enabling the analyst to probe into various user needs.

**Step 5: Conducting Chat Interviews with Personas:** The next step involves conducting simulated chat interviews with the selected personas. These interviews will utilize the capabilities of the LLM to engage in realistic, interactive dialogues with the personas, allowing the analyst to delve into the personas' hypothetical experiences, needs, and preferences. The LLM will generate responses in real-time, providing rich qualitative data that reflects the diversity and complexity of the user base.

This interview methodology offers several key benefits that enhance the effectiveness and flexibility of the research process. Firstly, the entire interview is recorded, allowing for subsequent revisits and reviews. This feature is particularly valuable in research, as it enables analysts to re-examine the responses at a later stage for deeper analysis or to clarify specific points. Moreover, the recorded nature of the interviews ensures that all responses are traceable, that ensures the integrity of the research, and for the verification of data and the tracking of insights back to their original sources. Finally, the flexibility in the duration of the interviews is a significant advantage. Analysts have the liberty to extend or shorten the interview time as needed, based on the depth of information required or the complexity of the topics being discussed.

**Step 6: User Story Generation Module:** The addition of a user story generation module will enable the LLM to craft user stories based on the simulated personas' attributes similar to the proposed tailored user story in [34]. This module will leverage the personas' diverse backgrounds and experiences to generate user stories that provide actionable insights for the analyst. The user stories will depict how different personas might interact with the software being developed, highlighting their specific needs, pain points, and the context of use.

## 6. Illustrative Case Studies

### 6.1 Global Entry Access System

Business Analyst 1 (BA1), a business analyst at SecurePass, is steering a groundbreaking project: the development of a Global Entry Access (GEA) system. The facial recognition system is tailored to meet the needs of a diverse array of international travelers. BA1's primary challenge is to ensure the system's inclusivity, catering to users across a spectrum of ethnicities, ages, genders, and physical abilities.

*Understanding the Recruitment Challenge:* BA1 recognizes the difficulty in recruiting participants who can provide a breadth of insights necessary for the GEA system. The system must not only recognize different skin tones but also account for age-related features, accommodate a range of gender identities, and be sensitive to cultural attire and physical disabilities. Given the global scope of the project, traditional recruitment methods were neither practical nor cost-effective.

*Harnessing D&I in AI:* To address this, BA1 turns to our proposed advanced D&I software. BA1 inputs critical diversity attributes that the GEA system must recognize:

- A broad range of skin tones to prevent ethnic bias.
- Ages extending from young adults to the elderly.
- Gender diversity including cisgender, transgender, and non-binary people.
- Physical features such as scars, prosthetics, or cultural headgear like hijabs.
- Considerations for users with disabilities, particularly visual impairments.

*Persona Generation:*

In a known-unknown situation like this, where the diversity attributes are known but their intersectionality as well as how they may interact with the GEA system is unknown, BA1 can effectively use the proposed tool. As an example below, through simulated interviews with these personas, BA1 uncovers specific system requirements. Some examples of the personas in this scenario can be:

- 'Amina', who wears a hijab, represents users who wear cultural head coverings.
- 'Lucas', a transgender man, who brings forward the need for sensitive gender recognition.
- 'Chen', an elderly individual with a prominent facial scar, highlighting the need for the system to accommodate unique facial features.
- 'Sophia', who is visually impaired, emphasizing the necessity for the system to be accessible to users with different abilities.

*Eliciting Requirements Through Personas:* Through simulated interviews with these personas, BA1 uncovers specific requirements:

- 'Amina' reveals the need for the system to identify users accurately without requiring the removal of cultural attire.
- 'Lucas' informs the design of gender-neutral algorithms that avoid misgendering users.
- 'Chen' contributes to the understanding that the system must discern individuals with distinctive facial features securely and conveniently.
- 'Sophia' highlights the importance of having audio feedback to guide visually impaired users during the recognition process.

The system can further generate user-stories for BA1 to select based on the context of the project and the D&I insights gathered. It can lead to further recruitment of other participants which may have only one specific diversity attribute. In the previous example. 'Chen' represents two attributes, age and facial scar. BA1 can request further personas to interview with focus on only one aspect of the diversity.

Influencing System Development: Informed by these insights, BA1 ensures that the GEA system's development includes diverse data samples for its learning algorithms, focuses on non-biased gender recognition, and incorporates essential accessibility features.

## 6.2 Adaptive Learning Platform

BA2, a business analyst at EdTech Innovations, is tasked with enhancing an adaptive learning platform intended for a global student base. The platform needs to cater to diverse learning styles and backgrounds, accommodating intersectional diversity attributes that are challenging to source in the real world.

*The Challenge of Intersectionality:* The adaptive learning platform aims to be accessible to students with various combinations of attributes that influence their learning experience. BA2 must consider intersectional diversity, accounting for attributes such as socio-economic background, educational disparities, cognitive and physical abilities, cultural contexts, language proficiencies, and gender identities. Finding participants who represent the intersection of these attributes poses a significant recruitment challenge, as these combinations of attributes can be rare and difficult to access.

*Utilizing D&I Software:*

While BA2 knows the types of diversity attributes to consider the unknown lies in how these intersecting attributes specifically impact the learning experience and what precise features the platform needs to effectively cater to them. BA2 can leverage D&I-focused persona software to generate single focused or intersectional personas, as well as ask the D&I software to recommend personas with attributes such as:

- Socio-economic backgrounds ranging from low-income rural areas to high-income urban settings.
- Educational experiences span from limited formal education to highly educated but potentially re-entering education.
- Cognitive diversity including neurodiverse conditions like ADHD and autism spectrum disorders.
- Physical abilities encompassing a spectrum from full mobility to various mobility impairments.
- Cultural backgrounds reflecting a global student population, including non-native English speakers.
- Gender identities across the spectrum, including non-binary and transgender students.

*Persona Creation and Requirement Elicitation:* The software can provide BA2 with complex personas such as:

- 'Miguel', a first-generation college student from a low-income, rural area with dyslexia.
- 'Priya', a non-binary individual from an urban high-income background, who is on the autism spectrum.
- 'Fatima', a wheelchair-using student from a war-torn region, re-entering education after a prolonged disruption.
- 'Ken', an elderly student with partial hearing, returning to education to learn new skills for a career change.

*Drawing Insights from Personas:* Interacting with these personas, BA2 can gather detailed requirements:

- 'Miguel' needs text-to-speech functionality to support his dyslexia and content that is sensitive to his socio-economic context.
- 'Priya' benefits from a customizable user interface to reduce sensory overload and non-gendered avatars that affirm their identity.
- 'Fatima' requires a platform accessible through assistive technologies and resources that consider the psychological impact of her experiences.
- 'Ken' highlights the need for closed captioning and sign language resources to facilitate his learning journey.

*Impact on Development:* These insights enable BA2 to guide the development team in creating an adaptive learning platform that features:

- Advanced text-to-speech engines and customizable interfaces.
- Content that is culturally and socio-economically inclusive.
- Full compatibility with various assistive technologies.
- Resources in multiple languages, with considerations for non-native speakers.

BA2's approach ensures that EdTech Innovations' platform is inclusive technology in education, designed to empower students from all walks of life. BA2's use of D&I tools to consider intersectional attributes exemplifies how technology can adapt to meet the complex, varied needs of a global user base.

## 7. Expected Results

The expected outcomes of this proposed system are multifaceted, encompassing benefits in terms of efficiency, cost-effectiveness, traceability, and inclusivity. Firstly, by leveraging a repository of AI-synthesized personas, the system is expected to significantly reduce the time and financial resources typically required for participant recruitment in traditional methods. This efficiency is a game changer, particularly for projects with limited budgets or stringent timelines. Secondly, the traceability aspect of the system is a key advantage. With AI-driven user stories derived from real-world data, every decision and development in AI systems can be traced back to specific, diverse user needs, enhancing accountability and transparency of the development process. The comprehensive nature of the repository ensures a high degree of inclusivity in AI system development. By encompassing a broad spectrum of real-world diversity, the system ensures that AI solutions are not just technically sound but are also deeply attuned to the varied and nuanced requirements of a diverse user base. Ou proposed vision is expected to streamline the development process but also to establish a new standard for inclusivity and ethical consideration in AI system development, making technology more equitable and accessible for all.

The work is currently in progress. We are amid developing our repository of personas and are actively engaged in experimental interviews with these personas. So far, the results from these experiments are aligning well with what we hoped to achieve. The feedback and insights we're gaining from these AI-generated persona interviews are proving invaluable, and they are shaping the direction and effectiveness of our work. Economically, it's proving to be a viable solution, utilizing our limited D&I resources efficiently without sidelining other important development areas. In terms of stakeholder representation, the system is shaping up to be exceptionally comprehensive. As we develop and interact with a diverse range of personas, it's evident that the system is considering the impact on various user groups, thereby ensuring inclusivity in design. Lastly, the aspect of traceability is becoming increasingly apparent. We're able to see how D&I concerns are being raised for design considerations. This transparency is not just a gain for accountability but is also shaping the system to be more responsive to the nuanced needs of all relevant stakeholders.

## 8. Conclusion

Operationalizing diversity and inclusion in the development of AI systems is not merely a theoretical ideal but a pragmatic necessity. The research presented here strives to bridge the gap between the aspirational ethics of AI and their practical application by providing a structured, multi-stage research and development approach to integrate D&I principles holistically within AI ecosystems. The proposed research vision presents a transformative approach to integrating D&I within the AI development lifecycle from the outset through the innovative use of personas. By developing a dynamic and rich persona repository that draws from a vast array of real-world data and utilizing AI to operationalize these personas, the framework aims to bring depth and authenticity to user representations. From the creation of a program interface that taps into a LLM API, to the generation of user stories that cater to a spectrum of diversity attributes, this research outlines a method that is not only technically feasible but also ethically sound. This vision sets the stage for a future where AI systems are developed with a conscientious commitment to reflecting and respecting the full diversity of society.